# Reasoning with Mass Distributions

**Rudolf Kruse**  **Detlef Nauck**  **Frank Klawonn**
Dept. of Computer Science
Technical University of Braunschweig
Bueltenweg 74 - 75
W-3300 Braunschweig, Germany

## Abstract

The concept of movable evidence masses that flow from supersets to subsets as specified by experts represents a suitable framework for reasoning under uncertainty. The mass flow is controlled by specialization matrices. New evidence is integrated into the frame of discernment by conditioning or revision (Dempster's rule of conditioning), for which special specialization matrices exist. Even some aspects of non-monotonic reasoning can be represented by certain specialization matrices.

## 1 INTRODUCTION

In this paper we present a suitable theoretical model for handling uncertainty, which is an important problem in the range of knowledge based systems. Uncertainty corresponds to the valuation of some datum, reflecting the faith or doubt in the respective source. So we have to deal with statements being not just simply true or false but with a validity which is a matter of degree. This is caused by the fact that the actual state of the world is not completely determined and we have to rely on a human experts subjective preferences among different possibilities.

Throughout this paper we will restrict ourselves to the treatment of subjective valuations of evidence which requires the use of belief functions measuring the credibility of information although our concept of specialization is very general and can be applied to probabilities as well as possibility measures.

Let $\Omega$, a finite nonempty set be our *frame of discernment*. We assume $\Omega$ to be a product space $\Omega^M \triangleq \Omega_1 \times ... \times \Omega_m$ with $m$ characteristics $X^{(1)} \in \Omega_1, ..., X^{(m)} \in \Omega_m$ where $\Omega_i$ ($i = 1,..., m$) is a finite nonempty set. The partial knowledge is encoded through evidence masses attached to subsets of $\Omega$. Specialization matrices quantify the flow of masses, the concept we prefer to Dempster's rule of conditioning (Shafer 1976).

A mass distribution is considered here as the condensed representation of a (possibly unknown) random set, for other semantics see (Kruse, Schwecke and Heinsohn 1991).

Section 2 provides an overview about mass distributions and belief functions. In section 3 we present our main concept: the flow of evidence masses given by a specialization matrix (Kruse and Schwecke 1990). In section 4 we consider specialization matrices which can be applied to conditioning and revision, and discuss certain aspects of non-monotonic reasoning.

## 2 REPRESENTING KNOWLEDGE WITH MASS DISTRIBUTIONS

Belief functions aim to model a human decision maker's subjective valuation of evidence. For this purpose we consider an inaccessible, finite probability space $\Theta$ of sensors or experts and a sample space $\Omega$ containing the possible events. The sensors or experts choose subsets of $\Omega$ which they believe to contain the actual state of the world. This means we consider multivalued mappings defined on a probability space, here called random sets (Matheron 1975). With respect to the probability distribution on the sensor space $\Theta$ one unit of "belief" which we conceive as movable "evidence mass" is distributed among the elements of $\Omega$, attributing a greater amount to the more likely elements (the elements chosen by the most or most reliable sensors or experts). That means a *mass distribution m* (basic probability assignment (Shafer 1976)) is specified, which is a mapping from $2^\Omega$ to the unit intervall.

**Definition 1:** *Each function* $m : 2^\Omega \to [0, 1]$ *is called a mass distribution, whenever*

(i)   $m(\emptyset) = 0$,

(ii) $\quad \sum_{A: A \subseteq \Omega} m(A) = 1$

hold.

The mass $m(A)$ is understood to be the measure of "belief" that is committed exactly to $A$ and corresponds to the support given to $A$ but not to any strict subset of $A$. Those sets $A$ with $m(A) > 0$ are called *focal elements*. To obtain the *total* measure of belief committed to some set $A$, we have to sum up the quantities $m(B)$ for all $B \subseteq A$.

**Definition 2:** *If $m$ is a mass distribution on $2^\Omega$, then the function* $\text{Bel}_m : 2^\Omega \to [0, 1]$,

$$\text{Bel}_m(A) \stackrel{d}{=} \sum_{B: B \subseteq A} m(B),$$

*is called the belief function induced by $m$.*

$\text{Bel}_m(A)$ represents the degree to which the actual evidence supports $A$, i.e. it measures the *credibility* of $A$. We are also able to calculate the degree to which the evidence fails to refute $A$, i.e. the degree to which $A$ remains *plausible*:

$$\text{PL}_m(A) \stackrel{d}{=} 1 - \text{Bel}_m(\bar{A})$$

$$= 1 - \sum_{B: B \subseteq \bar{A}} m(B) = \sum_{B: A \cap B \neq \emptyset} m(B).$$

We have

$\text{Bel}_m(A) \leq \text{PL}_m(A)$, and

$\text{Bel}_m(A) + \text{Bel}_m(\bar{A}) \leq 1$

for all $A \subseteq \Omega$.

To measure the evidence mass that can freely move to any element or subset of $A$ we use the concept of *commonality functions*. Let $m$ be a mass distribution defined on $2^\Omega$. The function

$$Q_m(A) \stackrel{d}{=} \sum_{B: A \subseteq B} m(B)$$

measures the evidence mass which is attached to supersets of $A$ and can move to $A$ or to any of its subsets. Obviously $Q_m(A) = 0$ indicates that there is no mass "above" $A$, i.e. $A$ cannot receive more evidence mass from its supersets.

To represent *total ignorance* about the domain under consideration, we set $m(\Omega) = 1$ and $m(A) = 0$ for all $A \neq \Omega$ and we obtain $\text{Bel}_m(\Omega) = 1$ and $\text{Bel}_m(A) = 0$ for all $A \neq \Omega$. This belief function is called the *vacuous belief function*. On the other hand setting $m(\{x_i\}) = p_i$, $x_i \in \Omega = \{x_1, \ldots, x_n\}$ and $m(A) = 0$ for all non-elementary sets $A$ leads to a *Bayesian belief function* or, in terms of the probability theory, a discrete probability distribution. We can imagine "belief" as partially movable evidence mass, where $m(A)$ is that amount of mass which can, in the light of new information, move to every subset of $A$ but not to sets with elements outside of $A$.

The concepts of *conditioning* and *revision* are based on this idea. When we obtain the information that "the truth" is within some set $E$ with certainty, all elements of $\bar{E}$ become impossible. The two concepts differ in their treatments of sets which have a nonempty intersection with $E$. Conditioning a mass distribution $m$ defined on $\Omega$ with respect to a set $E \subseteq \Omega$ means to neglect the evidence mass which is inconsistent with the new information. All masses not attached to subsets of $E$ are omitted and the remaining masses are normalized.

**Definition 3:** *Let $m$ be a mass distribution on $2^\Omega$ and $E$ be a subset of $\Omega$ with $\text{Bel}_m(E) > 0$. The mass distribution*

$$m(.|E): 2^\Omega \to [0, 1]; \quad m(A|E) \stackrel{d}{=} \begin{cases} \dfrac{m(A)}{\text{Bel}_m(E)} & \text{if } A \subseteq E \\ 0 & \text{otherwise} \end{cases}$$

*is called conditional[1] mass distribution.*

The concept of revision is directly based on the idea of partially movable evidence mass. All masses attached to subsets $A$ of $\Omega$ float to the sets $A \cap E$ after revision with respect to the set $E$.

**Definition 4:** *Let $m$ be a mass distribution on $2^\Omega$ and $E$ be a subset of $\Omega$ with $\text{Bel}_m(E) > 0$. The mass distribution*

$$m_E : 2^\Omega \to [0, 1];$$

$$m_E(A) \stackrel{d}{=} \begin{cases} \dfrac{\sum_{D: D \cap E = A} m(D)}{\text{PL}_m(E)} & \text{if } A \neq \emptyset \\ 0 & \text{otherwise} \end{cases}$$

*is called revised[2] mass distribution.*

Contrary to conditioning revision does not omit the evidence mass attached to sets lying just partially in $E$. Revising $m$ on $E$ yields the belief function

$$\text{Bel}_{m_E}(A) = \frac{\text{Bel}_m(A \cup \bar{B}) - \text{Bel}_m(\bar{B})}{1 - \text{Bel}_m(B)}, \quad A \subseteq \Omega$$

and the plausibility function

$$\text{PL}_{m_E}(A) = \frac{\text{PL}_m(A \cap B)}{\text{PL}_m(B)}, \quad A \subseteq \Omega.$$

Remembering our idea of experts or sensors choosing subsets of $\Omega$ the differences between the two concepts conditioning and revision can be made clear quite easily. Conditioning is a very strict treatment of experts whose valuations are inconsistent with the new information $E$. These experts are now considered as totally unreliable and the evidence mass distributed due to their statements has to be redistributed under the subsets $A \subseteq E$ chosen by the reliable experts.

---

[1] This concept is also called *strong conditioning* (Dubois and Prade 1986a) or *geometric conditioning*.

[2] This concept is also know as Dempster's rule of conditioning (Shafer 1976).

Revision induces a more optimistic treatment of the experts. The idea is that the valuations which are only partially inconsistent with the new information ($A \not\subseteq E$ but $A \cap E \neq \emptyset$) are now treated as if the expert meant $A \cap E$ and not $A$. The expert just was not able to express this situation because he had not enough information. So he is still considered to be reliable and the evidence mass attached to $A$ flows completely to the intersection with $E$. Only those experts whose valuations are totally inconsistent with $E$ are treated as in the case of conditioning.

## 3 THE CONCEPT OF SPECIALIZATION

In order to compare different frames of discernment we introduce the notion of a refinement (Shafer 1976).

**Definition 5:** *A set $\Omega'$ is a refinement of $\Omega$ if there is a mapping $\hat{\Pi} : 2^\Omega \to 2^{\Omega'}$ such that*

(i) $\hat{\Pi}(\{x\}) \neq \emptyset$ for all $x \in \Omega$,

(ii) $\hat{\Pi}(\{x\}) \cap \hat{\Pi}(\{x'\}) = \emptyset$, if $x \neq x'$,

(iii) $\bigcup \left\{ \hat{\Pi}(\{x\}) \mid x \in \Omega \right\} = \Omega'$ and

(iv) $\hat{\Pi}(A) = \bigcup \left\{ \hat{\Pi}(\{x\}) \mid x \in A \right\}$.

$\hat{\Pi}$ is called a refinement mapping. If such a mapping exists, the sets $\Omega$ and $\Omega'$ are compatible, where the refined space $\Omega'$ is able to carry more information than its quotient space $\Omega$. In order to decide for each $\omega \in \Omega$ whether information concerning some set $A' \subseteq \Omega'$ may be of relevance for the valuation of $\omega$ or not we define the mapping $\Pi$.

**Definition 6:** *Let $\Omega'$ be a refinement of $\Omega$ where $\hat{\Pi} : 2^\Omega \to 2^{\Omega'}$ is the respective refinement mapping. The mapping*

$$\Pi : 2^{\Omega'} \to 2^\Omega, \quad \Pi(A') \stackrel{d}{=} \left\{ \omega \in \Omega \mid \hat{\Pi}(\{\omega\}) \cap A' \neq \emptyset \right\}$$

*is called the outer reduction induced by $\hat{\Pi}$.*

$\Pi(A')$ contains those $\omega \in \Omega$ which have one or more elements $\omega' \in \hat{\Pi}(\{\omega\})$ within $A'$. Note that $\Pi$ essentially is a projection that attaches to each element $\omega' \in \Omega$ that element $\omega$ with $\omega' \in \hat{\Pi}(\{\omega\})$. The projection of a mass distribution $m'$ defined on $2^{\Omega'}$ can be obtained by

$$\Pi(m') : 2^\Omega \to [0,1]; \quad \Pi(m')(A) \stackrel{d}{=} \sum_{\substack{A' \subseteq \Omega'; \\ \Pi(A')=A}} m'(A').$$

If there is a mass distribution $m'$ defined on $2^{\Omega'}$ and a projection $\Pi(m')$ of $m'$ on $2^\Omega$, then $m'$ is a refinement of $\Pi(m')$. The formulation of a mass distribution $m$ on $\Omega$ in terms of the refined space $\Omega'$ is defined by

$$\hat{\Pi}(m) : 2^{\Omega'} \to [0,1];$$

$$\hat{\Pi}(m)(A') \stackrel{d}{=} \begin{cases} m(A), & \text{if } A' = \hat{\Pi}(A), \\ 0 & \text{otherwise} \end{cases}$$

and is denoted as the *vacuous extension* of $m$. From the definition it is obvious, that each vacuous extension of a mass distribution is its refinement. In contrast to the projection which generally means a loss of information, the vacuous extension preserves the information borne by the original mass distribution.

The main issue of this chapter is to define the concept of specialization. The intuitive idea of a specialization is the projection of a revision.

**Definition 7:** *Let $s$, $t$ be two mass distributions defined on $2^\Omega$. We call $s$ a specialization of $t$ ($s \sqsubset t$), if and only if there are two mass distributions $s'$ and $t'$ on a refinement $\Omega'$ of $\Omega$ where $s'$ and $t'$ are refinements of $s$ and $t$, respectively, and if there is an event $E' \subseteq \Omega'$ such that*
$$s'(B') = t'_{E'}(B')$$
*holds for each $B' \subseteq \Omega'$.*

This definition tells us that we will get all specializations of a given mass distribution on $\Omega$ by considering all possible revisions in a refined space $\Omega'$. Relating now the concept of specialization with Dempster's rule of combination we can see, that specialization is bound to the idea of *updating* and not to *aggregation*. Dempster's rule combines two mass distributions (basic probability assignments) which are defined on the same sample space but based on different bodies of evidence. This is a concept of aggregating different expert views.

The change from a mass distribution $m$ to a specialization of $m$ is a different concept, and it is due to an updating of the refinement of $m$ in a refinement of the sample space $\Omega$. We use revision as the updating rule which causes a change of data in the refined space. Those observations $A$ of the experts which are not completely covered by the new evidence $E$ are changed to become $A \cap E$ instead without loosing any evidence mass.

In addition to the definition above the following theorem gives two equivalent characterizations of the specialization relationship. The first one allows to check easily whether $s \sqsubset t$ is valid or not. The second one reflects our intuitive idea of floating evidence masses describing the flow of the mass $t(A)$ onto the subsets of $A$.

**Theorem 1:** *Let $s, t$ be two mass distributions on $\Omega$. The following three statements are equivalent:*

(i) $s \sqsubset t$,

(ii) $\forall A \subseteq \Omega : ( Q_t(A) = 0 \Rightarrow Q_s(A) = 0 )$,

(iii) For every $A \subseteq \Omega$ there are functions
$h_A : 2^\Omega \to [0,1]$ such that

a) $\sum_{B:B \subseteq \Omega} h_A(B) = t(A)$,

b) $h_A(B) \neq 0 \Rightarrow B \subseteq A$, for all $B \subseteq \Omega$, and

c) $s(B) = \dfrac{\sum_{A:A \subseteq \Omega} h_A(B)}{1 - \sum_{A:A \subseteq \Omega} h_A(\emptyset)}$ for all $\emptyset \neq B \subseteq \Omega$.

$h_A(B)$ specifies that amount of "belief" comitted to $A$ that in the course of refining $m$ to $m'$ floats to the set $B$. Condition (iii.a) of Theorem 1 assures that no evidence mass is lost, condition (iii.b) requires that the masses flow only to subsets. Those masses floating to the empty set represent partial contradictions, thus have to be neglected and the remaining portions have to be normalized as pointed out in condition (iii.c).

The normalization in condition (iii.c) is due to our treatment of experts whose observations are totally inconsistent with the new evidence (see sect. 2). They are now considered to be unreliable and so the evidence mass bound to their observations has to be redistributed under the consistent observations. Note that we also use a *closed world assumption*. Smets (Smets 1988) considers an *open world assumption* and allows the empty set to bear evidence mass. In this case there is no normalization of the remaining masses because the evidence mass on the empty set is supposed to indicate the belief that the actual state of the world cannot be represented in the chosen frame of discernment. Our perception of the empty set is a different one. The evidence mass that flows to the empty set indicates from our point of view the inconsistency of expert observations at the beginning of the updating process and is not characterizing the current situation. So a normalization has to be made because we don't want to weaken the belief in the consistent observations. Using an open world assumption means that an expert cannot be wrong in spite of inconsistencies due to new information. From our point of view inconsistency arises because of errors made by some of the experts.

A similar concept to the specialization relation is the idea of a containment of "bodies of evidence" introduced in (Yager 1986). A body of evidence is a pair $(F,m)$, where $m$ is a mass distribution defined on $\Omega$ and $F$ contains the focal elements of $m$. A definition of "strong inclusion" can be found in (Dubois and Prade 1986b):

$(F,m) \prec (F',m')$ if and only if

(i) $\forall B \in F, \exists A' \in F', B \subseteq A'$

(ii) $\forall A' \in F', \exists B \in F, B \subseteq A'$

(iii) There exist $W_{BA'} \in [0,1]$, for all $B, A'$ such that
$W_{BA'} > 0 \Rightarrow B \subseteq A', \sum_{A',B} W_{BA'} = 1$, and

$\forall B \in F, m(B) = \sum_{A':B \subseteq A'} W_{BA'}$,

$\forall A' \in F', m'(A') = \sum_{B:B \subseteq A'} W_{BA'}$.

Specialization is more general than strong inclusion. We have $(F,m) \prec (F', m') \Rightarrow m \sqsubset m'$ but not vice versa. The $W_{BA'}$ are identical to the values $h_A(B)$, but there is no normalization. From considering the definition above and our idea of floating evidence masses, it is obvious that in the case of strong inclusion there is no mass flow to the empty set and that no mass is lost ($\sum W_{BA'} = 1$), so a normalization is not necessary.

## 4 SPECIALIZATION MATRICES

In order to compute a specialization of a mass distribution $m$ we characterize $m$ as a vector and the respective specialization-relation by a matrix $V : 2^\Omega \times 2^\Omega \to [0, 1]$ and obtain the more specific mass distribution $m'$ by "multiplying" the vector $m$ with the matrix $V$. In the following we use square brackets to indicate that we conceive the respective functions as vectors or matrices.

**Definition 8:** *Let $\Omega$ be the frame of discernment.*
(i) *A matrix $V : 2^\Omega \times 2^\Omega \to [0, 1]$ is called a specialization matrix, if and only if*

(a) $\sum_{B:B \subseteq \Omega} V[A,B] = 1$ for all $A \subseteq \Omega$

(b) $B \not\subseteq A \Rightarrow V[A,B] = 0$.

(ii) *Let $V$ be a specialization matrix and let $m$ be a mass distribution on $2^\Omega$. If*

$$c \stackrel{d}{=} \sum_{A:A \subseteq \Omega} \sum_{B:B \neq \emptyset} m[A] \cdot V[A,B] > 0$$

*then the mass distribution $m \odot V$ is defined by*

$$(m \odot V)[B] \stackrel{d}{=} \begin{cases} \frac{1}{c} \cdot \sum_{A: A \subseteq \Omega} m[A] \cdot V[A,B] & \text{if } B \neq \emptyset \\ 0 & \text{otherwise} \end{cases}$$

for all $B \subseteq \Omega$.

In contrast to the mass flow functions $h_A$, $A \subseteq \Omega$, specialization matrices do not assign absolute portions but relative amounts of mass.

**Theorem 2:** *Let $m$ and $m'$ be two mass distributions defined on $2^\Omega$. We have*

$$m' \sqsubset m \Leftrightarrow \exists V: m' = m \odot V,$$

*where $V$ is a specialization matrix.*

The processes of conditioning and revision, i.e. the change from a mass distribution $m$ to the conditional mass distribution $m(\cdot|E)$ or to the revised mass distribution $m_E$ respectively, are special cases of specialization and can therefore be described by special specialization matrices.

Recall that conditioning with respect to the set $E \subseteq \Omega$ means that those masses bound to sets $A \subseteq E$ remain where they are, while those bound to sets $A \not\subseteq E$ have to be neglected.

**Definition 9:** *Let $\Omega$ be the frame of discernment and let $E \subseteq \Omega$ be a non-empty set. The conditional matrix $C(E) : 2^\Omega \times 2^\Omega \to [0, 1]$ is defined by*

$$C(E)[A,B] \stackrel{d}{=} \begin{cases} 1 & \text{if } A \not\subseteq E \text{ and } B = \emptyset \\ 1 & \text{if } A \subseteq E \text{ and } B = A \\ 0 & \text{otherwise} \end{cases}$$

We obtain $m(\cdot|E) = m \odot C(E)$.

Revision with respect to the set $E$ means that the masses attached to sets $A \neq \emptyset$ float to $A \cap E$. Masses attached to sets with $A \cap E = \emptyset$ have to be neglected since they represent (partial) contradictions of the information $E$ and the mass distribution $m$.

**Definition 10:** *Let $\Omega$ be the frame of discernment and let $E \subseteq \Omega$ be a non-empty set. The revision matrix $R(E) : 2^\Omega \times 2^\Omega \to [0, 1]$ is defined by*

$$R(E)[A,B] \stackrel{d}{=} \begin{cases} 1 & \text{if } B = A \cap E \\ 0 & \text{otherwise} \end{cases}$$

We obtain $m_E = m \odot R(E)$.

The use of specialization matrices leads to a new interesting concept. Some specialization matrix $V$ represents a piece of "structural knowledge". Multiplying a mass distribution $m$ with $V$ means to split the evidence masses in the light of knowledge encoded by $V$. A rather strict requirement is that the "application" of $V$ to a more specific mass distribution $m'$ should yield a more specific result.

**Definition 11:** *Let $V : 2^\Omega \times 2^\Omega \to [0, 1]$ be a specialization matrix. $V$ is called monotonic, if and only if*

$$s \sqsubset t \Rightarrow s \odot V \sqsubset t \odot V$$

*holds for all mass distribution $s$, $t : 2^\Omega \to [0, 1]$.*

The next theorem provides a simple possibility to check whether a given specialization matrix is monotonic or not. It relies on a test, if there is no such set $A$ whose mass flow is completely "outrun" by one of its supersets mass flow.

**Theorem 3:** *Let $V : 2^\Omega \times 2^\Omega \to [0, 1]$ be a specialization matrix. $V$ is monotonic, if and only if for all sets $A$, $B \subseteq \Omega$ with $V[A,B] > 0$, and for all $C \supseteq A$ there is a set $D \supseteq B$ with $V[C,D] > 0$.*

**Theorem 4:** *Let $s,t$ be two mass distributions defined on $2^\Omega$ and $s \sqsubset t$. Then there is always a specialization matrix $V : 2^\Omega \times 2^\Omega \to [0, 1]$ and $V$ is monotonic, such that $s = t \odot V$.*

We want to show in the sequel that also aspects of non-monotonic reasoning can be handled with specialization matrices. From Theorem 3 it is clear that a specialization matrix $V$ is *non-monotonic*, if there exist sets $B \subseteq A \subseteq C$ such that there is a mass flow from $A$ to $B$ and no mass flow from $C$ to supersets of $B$.

First we want to compare non-monotonic specialization matrices with Yager's non-monotonic compatibility relations (Yager 1988). Yager defines a (type II) compatibility relation on two sets $X$ and $Y$ as a relation $R$ on $2^{X'} \times Y$ such that for each $T \in 2^{X'}$ there exists at least one $y \in Y$ such that $(T,y) \in R$, where $2^{X'}$ is the power set of $X$ minus the empty set. $R(T,y)$ implies that $(x,y)$, for all $x \in T$, are possible states of the world.

Let $W = \{y | R(T,y)\}$ be the subset of $Y$ that contains the $y \in Y$ which are related to any $x \in T$. $W$ is called the "associated set" in $Y$ of $T$, denoted $T \to W$. A compatibility relation $R$ is called "irregular" if there exists a triple $T_1 \to W_1$, $T_2 \to W_2$ and $T_3 \to W_3$ with $T_3 = T_1 \cup T_2$ such that $W_3$ is strictly contained in $W_1 \cup W_2$, $W_3 \subset W_1 \cup W_2$. Yager has proven that every irregular (type II) compatibility relation is *non-montonic*. That means if we have two mass distributions $s$, $t$ and we have $s \prec t$ (strong inclusion) this does not imply $s \circ R \prec t \circ R$.

Because the concept of specialization matrices is more general than compatibility relations, a non-monotonic compatibility relation $R$ can be easily expressed with a non-monotonic specialization matrix. Let $S$ be a subset of $X \times Y$, let $D_S = \{x \mid \exists y, (x,y) \in S, R(x,y)\}$, and let $W_{D_S}$ be the associated set of $D_S$. A (type II) compatibility relation R can be expressed with a specialization matrix $V_R: 2^{X \times Y} \times 2^{X \times Y}$ with

$$V_R[S,S'] = \begin{cases} 1, & \text{if } S' = S \cap \{D_S \times W_{D_S}\} \\ 0, & \text{otherwise} \end{cases}$$

If the relation $R$ is non-monotonic, the same is true for the specialization matrix $V_R$. If we express any (type II) compatibility relation $R$ with a specialization matrix $V_R$, and $V_R$ is non-monotonic, the same holds for $R$.

Now let us take a look at the well known example of the bird Tweety who is not able to fly because he is a penguin. Let $\Omega = \Omega_1 \times \Omega_2$ be our frame of discernemt where $\Omega_1 = \{birds, fish\}$ and $\Omega_2 = \{fly, not\ fly\}$. Now the rule "All birds fly" can be expressed by a specialization matrix $V$ with

$$V[A,B] \stackrel{d}{=} \begin{cases} 1 & \text{if } B = A - \{(birds, not\ fly)\} \\ 0 & \text{otherwise} \end{cases}$$

The rule "Penguins don't fly" can only be represented in a refined space, e.g. $\Omega' = \Omega_1' \times \Omega_2$, where $\Omega_1' = \{eagles, penguins, fish\}$. In our refined space the two (partially contradicting) rules "All birds fly" and "Penguins don't fly" are expressed by the following specialization matrix $V'$.

$$V'[A,B] \stackrel{d}{=} \begin{cases} 1 & \text{if } A \supseteq \{eagles, penguins\} \times \{not\ fly\} \\ & := H \text{ and } B = A - H, \\ 1 & \text{if } (penguins, fly) \in A \text{ and } B \\ & = A - \{(penguins, fly)\}, \\ 1 & \text{if } A \not\supseteq H \text{ and } (penguins, fly) \notin A, \\ 0 & \text{otherwise} \end{cases}$$

The two rules force the mass attached to the set $C = \{eagles, penguins\} \times \{fly, not\ fly\}$ to float to the set $D = \{eagles, penguins\} \times \{fly\}$ and the masses attached to $A = \{penguins\} \times \{fly, not\ fly\}$ to $B = \{penguin\} \times \{not\ fly\}$. We have $C \supseteq A$ but $D \not\supseteq B$. That means the specialization matrix $V'$ is non-monotonic.

## 5   CONCLUSIONS

With the calculus of mass distributions we presented a suitable theoretical tool for reasoning under uncertainty. We showed that the flow of evidence masses can be conveniently handled by specialization matrices. For the concepts of conditioning and revision (Dempster's rule of conditioning) there exist special specialization matrices. We also demonstrated that certain aspects of non-monotonic reasoning, especially partially contradicting statements can be expressed by non-monotonic specialization matrices. In cooperation with Dornier GmbH the method of reasoning with mass distributions was implemented on a TI-Explorer under KEE.